\documentclass[11pt]{article}
\usepackage{acl-hlt2011}
\usepackage{times}
\usepackage{latexsym}
\usepackage{amsmath}
\usepackage{multirow}
\usepackage{url}
\usepackage{graphics}
 \usepackage{graphicx}

\newenvironment{myitemize}{
\vspace{-0.3\baselineskip}
\begin{itemize}
\setlength{\topsep}{0pt}
\setlength{\itemsep}{0pt}
\setlength{\parskip}{0pt}
\setlength{\parsep}{0pt}
\setlength{\partopsep}{0pt}
}{
\end{itemize}
\vspace{-0.2\baselineskip}}
\title{Even the Abstract have Colour:\\ Consensus in Word--Colour Associations}

\author{Saif M. Mohammad \\
	Institute for Information Technology\\
    National Research Council Canada.\\
    Ottawa, Ontario, Canada, K1A 0R6\\
    {\tt saif.mohammad@nrc-cnrc.gc.ca}
}
\date{}

\begin{document}
\maketitle
\begin{abstract}
Colour is a key component in the successful dissemination of
information.  Since many real-world concepts are associated with
colour, for example {\it danger} with red, linguistic information is
often complemented with the use of appropriate colours in information
visualization and product marketing.  Yet, there is no comprehensive
resource that captures concept--colour associations.  We present a
method to create a large word--colour association lexicon by
crowdsourcing.  A word-choice question was used to obtain sense-level
annotations and to ensure data quality.  We focus especially on
abstract concepts and emotions to show that even  they tend to have
strong colour associations.  Thus, using the right colours can not
only improve semantic coherence, but also inspire the desired
emotional response.

% We show that more than 30\% of the terms have a strong colour association.
% There is no correlation between imageability and colour association,
% whereas emotions have  strong colour associations. 
% We also found that preference for colours has a remarkable correlation with
% the order in which colour terms appeared in language.
\end{abstract}

\section{Introduction}
% Much of the research in the overlap of language, cognition, and colour is on 
% proving or disproving the idea that language effects cognition (Brown and Lenneberg, 1954; Ratner, 1989 and Bornstein, 1985, respectively).
% However, there is a more practical aspect to the relation between language, colour, and cognition too.

Colour is a vital component in the successful delivery of information, whether it is
in marketing a commercial product \cite{SableA10}, in web design \cite{Meier88,Pribadi90}, or in information visualization  \cite{Christ75,CardMS99}.
% This is because
Since real-world concepts have associations with certain colour categories
(for example, {\it danger} with red, and {\it softness} with pink),
complementing linguistic and non-linguistic information with appropriate colours has a number of benefits, including:
(1) strengthening the message (improving semantic coherence), 
(2) easing cognitive load on the receiver,
(3) conveying the message quickly, and
(4) evoking the desired emotional response.
Consider, for example, the use of red in stop signs. % In North America, stop signs % have ``STOP" written on them, but they 
% are coloured red. Red has an association with danger, and so strengthens the information
% intended to be delivered by the sign. 
Drivers are able to recognize the sign faster, and it evokes a subliminal emotion pertaining to possible danger, which is entirely appropriate in the context.
% for a stop sign whose purpose is to avoid accidents.
The use of red to show areas of high crime rate in a visualization is another example of good use of colour to draw emotional response.
On the other hand, improper use of colour can be more detrimental to understanding than using no colour \cite{Marcus82,Meier88}.

% In a visualization of the effect of climate change on glaciers, snow sheets are best shown in white.

% Using the right colours is crucial in marketing a commercial product, in web design,
% For example, pinks are used with products targeting young girls. colours are also important 
% and in information visualization. 
% Visualizations are a graphical representation intended to give insights into data. The data pertains to different concepts, and 
% Using the right colours to represent different concepts will make it easier for the consumer to follow the information.
% For example, 
% Appropriate colours can also be used to convey the appropriate emotional effect a visualization or website is intended to evoke. 
% in a visualization of electoral results of USA, blue is better suited to represent Democrats and red for Republicans.

A word has strong association with a colour when the colour is a salient feature of the concept
 the word refers to, or because the word is related to a such a concept.
 Many concept--colour associations, such as {\it swan} with white and {\it vegetables} with green, involve
 physical entities.
However, even abstract notions and emotions
may have colour associations ({\it honesty}--white, {\it danger}--red, {\it joy}--yellow, {\it anger}--red).
Further, many associations are culture-specific \cite{Gage99,Chen05}. For example, {\it prosperity} is associated with
red in much of Asia.

% Some concept--colour associations, such as {\it swan} with white and {\it vegetables} with green, may be easy to recall,
% but there are innumerable concepts in the world. % and many of them may have associations with colours.
% % The associated concepts are not restricted to physical entities such as snow and plants,
% % but can also refer to 
% Even abstract notions, such as {\it honesty} and {\it danger}, and emotions
% such as {\it joy} and {\it anger}, may have colour associations.
% % In fact, there are far more concepts in the world than there are colour categories, which
% % means that the same colour may be associated with many concepts. For example, red
% % is associated with danger, passion, and republicans, to name a few.
% Further, many associations are culture-specific \cite{Gage99,Chen05}. For example, red is associated with
% good fortune in much of Asia.
% Not every concept is associated with a colour, and different concepts may be associated
% to colours to a different degree. 

Unfortunately, there exists no lexicon with any significant coverage that captures these concept--colour associations,
and a number of questions remain unanswered, such as, the  extent to which humans agree with each other on these associations,
and whether physical concepts are more likely to have a colour association than abstract ones.

In this paper, we describe how we created a large word--colour lexicon by crowdsourcing with effective quality control measures (Section 3),
as well as experiments and analyses to show that:

% We propose a method to determine the non-abstractness of thesaurus categories using 
 \begin{myitemize}
 \item More than 30\% of the terms % and more than blah\% of thesaurus categories 
 have a strong colour association (Sections 4). 
% \item There is a fair amount of agreement among humans for word--colour associations (Section 4).
% \item Frequencies of colour choice in associations follows the same order in which colour terms ﬁrst appeared in language.
\item  % We use the groupings of words in a thesaurus to show that even 
% Even thesaurus groupings of related words have strong colour associations (Section 5).
About 33\% of thesaurus categories have strong colour associations (Section 5).
% \item There is a definite order of preference for colours. If the colours are ranked in decreasing order of the number of times they were picked to be associated with a word,
% we arrive at the same order in which colour terms first appeared in most of the languages (Section 4).
% \item Even though it may seem that a larger percentage of concrete and imageable terms may have strong colour associations,
% \item There is very little, if any, correlation between the imageability of terms and their tendency to have an association with colour. In other words, 
\item Abstract terms have colour associations almost as often as physical entities do (Section 6).
\item There is a strong association between different emotions and colours (Section 7).
\end{myitemize}
% \noindent Thus, linguistic or non-linguistic representations of real-world concepts can often be shown in tandem with colours to not only improve semantic coherence, but also
% to inspire the appropriate psychological response.
% Frequencies of colour choice in associations follows the same order in which colour terms ﬁrst appeared in language.
\noindent Thus, using the right colours can not only improve
semantic coherence, but also inspire the desired emotional
response.

% -- Imagability\\
% -- Agreement\\
% - - - Art\\
% web design

\section{Related Work}
 The relation between language and cognition has received considerable attention
 over the years, mainly on answering whether language impacts thought,
 and if so, to what extent. Experiments with colour categories have been used both to show
 that language has an effect on thought \cite{BrownL54,Ratner89}
 and that it does not \cite{Bornstein85}. However, that line of work does not
explicitly deal with word--colour associations. 
In fact, we did not find any other academic work that gathered large word--colour associations. There is, however,
a commercial endeavor---Cymbolism\footnote{http://www.cymbolism.com/about}.

Child et al.\@ \shortcite{Child68}, Ou et al.\@ \shortcite{Ou11}, and others show that people of different ages and genders
have different colour preferences. (See also the online study by Joe Hallock\footnote{http://www.joehallock.com/edu/COM498/preferences.html}.)
In this work, we are interested in identifying words that have a strong association with a colour
due to their meaning; associations that are not affected by age and gender preferences.

% Luscher \shortcite{Luscher69} argued that there is a correlation between
% personal preference for certain colours and one's internal psychological state.
There is substantial work on inferring the emotions evoked by colour \cite{Luscher69,Kaya04}.
% In Section 7, we determine the association of {\it emotional words} with colours.
Strapparava and Ozbal \shortcite{StrapparavaO10} compute corpus-based semantic similarity between emotions and colours.
% We combine a word--colour and a word--emotion lexicon to determine the association between 
% emotion words and colours.
 We combine a word--colour and a word--emotion lexicon to determine the association between
  emotion words and colours.

% We investigate the correlation of emotions with colours in our work.
% proposed a test wherein a person is asked to
% rank coloured cards in order of preference. He argued that the order of these colours provides
% insight into the psychological state of the person. 
% Since then there have been both proponents () and detractors () of the theory.
% In our work, we will simply show that some thesaurus categories and emotions have strong colour associations.
% We do not argue that preference for certain colours is indicative of associated psycological states.
Berlin and Kay \shortcite{BerlinK69}, and later Kay and Maffi \shortcite{KayM99}, showed that often colour
terms appeared in languages in certain groups. If a language has only two colour terms, then they are white and black. If a language has three colour terms, then they tend to be white, black, and red.
Such groupings are seen for up to eleven colours, and based on these groupings, colours can
be ranked as follows:
% Below is the most common order (earlier to later):
% We propose a method to determine the non-abstractness of thesaurus categories using 
 \begin{quote}
1. white, 2. black, 3. red, 4. green, 5. yellow, 6. blue, 7. brown, 8. pink, 9. purple, 10. orange, 11. grey \hspace{27mm} (1)
\end{quote}
\noindent There are  hundreds of different words for colours.\footnote{See http://en.wikipedia.org/wiki/List\_of\_colors}
 To make our task feasible, % we needed to choose a relatively small list of basic colours. 
 we chose to use the eleven basic colour words of Berlin and Kay (1969).
% We used these as the source pool of colours in our annotations.

The MRC Psycholinguistic Database \cite{Coltheart81} has, among other information, the {\it imageability
ratings} for 9240 words.\footnote{http://www.psy.uwa.edu.au/mrcdatabase/uwa\_mrc.htm}
The imageability rating is a score given by human judges that reflects
how easy it is to visualize the concept.
It is a scale from 100 (very hard to visualize) to 700 (very easy to visualize).
We use the ratings  in our experiments to determine whether there is a correlation between imageability and
strength of colour association.

\begin{table*}[]
\centering
% \resizebox{\textwidth}{!}{
 {\small
\begin{tabular}{l rrrr rrrr rrr}
\hline
      %   \multicolumn{11}{c}{\bf colour} \\
            & white &black &red &green &yellow &blue &brown &pink &purple &orange &grey\\
\hline
            overall     &11.9       &12.2       &11.7       &12.0       &11.0       &9.4        &9.6        &8.6        &4.2  &4.2 &4.6\\
            voted   &22.7    &18.4       &13.4       &12.1       &10.0       &6.4        &6.3        &5.3        &2.1        &1.5 &1.3\\
\hline
\end{tabular}
}
\vspace*{-1mm}
\caption{Percentage of terms marked as being associated with each colour.}
\label{tab:col votes}
\vspace*{-1mm}
\end{table*}

\section{Crowdsourcing}
We used the {\it Macquarie Thesaurus} \cite{Bernard86} as the source for terms to be annotated by people on Mechanical Turk.\footnote{Mechanical Turk: www.mturk.com}
Thesauri, such as the {\it Roget's} and {\it Macquarie}, group related words into categories. These categories can be thought of as coarse senses \cite{Yarowsky92,MohammadH06b}.
If a word is ambiguous, then it is listed in more than one category.
Since we were additionally interested in determining colour signatures for emotions (Section 7),
we chose to annotate all of the 10,170 word--sense pairs that Mohammad and Turney \shortcite{MohammadT10} used to create their word--emotion lexicon.
% Since the {\it Roget Thesaurus} is available freely in the public domain, it allows us
% to distribute our emotion lexicon without the burden of restrictive licenses. 
% We chose only those words that occurred more than
% 120,000 times in the Google n-gram corpus.\footnote{The frequency threshold of 120,000 is arbitrary.
% We wanted to annotate the most frequent words first. The Google n-gram corpus is available through the Linguistic Data Consortium.}
% Thesauri such as the Roget and Macquarie divide the vocabulary into about a thousand coarse categories. Words in a category are closely related, and an ambiguous word may be listed
% in more than one category. These categories can be thought of as its coarse senses. Like Mohammad and Turney \shortcite{MohammadT10}, our emotion
% lexicon has entries at sense level. We bias the annotator towards a particular sense of the target word by presenting a related word from the
% target thesaurus category (target sense).
Below is an example questionnaire:

% \rule{7.6cm}{0.4mm}

\vspace*{1mm}
{\small
\indent Q1. Which word is closest in meaning  % (most related) 
to {\it sleep}?

% \vspace*{-1mm}
\begin{minipage}[t]{4mm}
\end{minipage}
\begin{minipage}[t]{1.7cm}
 \begin{myitemize}
\item {\it car}
\end{myitemize}
\end{minipage}
\begin{minipage}[t]{1.9cm}
 \begin{myitemize}
\item {\it tree}
\end{myitemize}
\end{minipage}
\begin{minipage}[t]{1.7cm}
 \begin{myitemize}
\item {\it nap}
\end{myitemize}
\end{minipage}
\begin{minipage}[t]{1.6cm}
\vspace*{-0.7mm}
 \begin{myitemize}
\item {\it olive}\\
\end{myitemize}
\end{minipage}

\vspace*{-2mm}
Q2. What colour is associated with {\it sleep}?

\vspace*{-1mm}
 \begin{minipage}[t]{4mm}
 \end{minipage}
\begin{minipage}[t]{1.7cm}
 \begin{myitemize}
 \item black
 \item blue
 \item brown
\end{myitemize}
\end{minipage}
\begin{minipage}[t]{1.8cm}
 \begin{myitemize}
 \item green
 \item grey
 \item orange
\end{myitemize}
\end{minipage}
\begin{minipage}[t]{1.7cm}
 \begin{myitemize}
 \item purple
 \item pink
 \item red
\end{myitemize}
\end{minipage}
\begin{minipage}[t]{1.9cm}
 \begin{myitemize}
 \item white
 \item yellow\\
\end{myitemize}
\end{minipage}
}
% \indent \rule{7.6cm}{0.4mm}

\vspace*{1mm}
\noindent 
Q1 is a word choice question generated automatically by taking a near-synonym
from the thesaurus and random distractors. 
% If a word has multiple senses, that is, it is listed in more than one thesaurus categor
If an annotator answers this question incorrectly,
then we discard information from both Q1 and Q2. 
The near-synonym also guides the annotator to the desired sense of the word.
Further, it
encourages the annotator to think clearly about the target word's meaning; we believe this improves the quality of the annotations in Q2.

The colour options in Q2 were presented in random order.
We do not provide a ``not associated with any colour" option to encourage
colour selection even if the association is weak. If there is no association
between a word and a colour, then we expect low agreement for that term.
We requested annotations from five different people for each term.
% question makes the annotator think about the meaning of the target word, which we think helps in getting better annotations for Q2.

The annotators on Mechanical Turk, by design, are anonymous. 
% We do not have their age and gender information. 
However, we requested annotations from US residents only.
%  The survey was approved by the ethics board at the authors' institution.

% Before going live, the survey was approved by the ethics committee
% at the National Research Council Canada.

% \section{Running on MTurk and post-processing}

\section{Word--Colour Association}

About 10\% of the annotations had an incorrect answer to Q1. Since, for these instances,
the annotator did not know the meaning of the target word, we discarded the corresponding
colour association response.
Terms with less than three valid annotations were discarded from further analysis. 
Each of the remaining terms has, on average, 4.45 distinct annotations.
% For 74.4\% of the those instances, all five annotators agreed on whether a term is associated
% with a particular emotion or not. For 16.9\% of the instances four out of five people agreed with each other,
% and for 8.5\% of the instances there was a three--two split.
The information from multiple annotators % for a particular term 
was combined
by taking the majority vote, resulting in a lexicon with 8,813 entries.
% The lexicon has 8,813 entries. % (we will increase this to about 15,000 in the near future).
Each entry contains a unique word--synonym pair, majority voted colour(s), and
a confidence score---number of votes for the colour
/ number of total votes.
(For the analyses in Sections 5, 6, and 7, ties were broken by picking one colour at random.) 
A separate version of the lexicon that includes entries for all of the valid annotations by each of the annotators
is also available.\footnote{Please contact the author to obtain a copy of the lexicon.}

The first row in Table \ref{tab:col votes} shows the percentage of times different colours were
associated with the target term. % The first row shows raw percentages.
The second row shows percentages after taking a majority vote of the annotators.
% For each term, the colour that gets the majority votes (from the 3 to 5 annotators
% for that HIT) is chosen
% as the true colour associated with the target term, and the ``voted" row
% of Table \ref{tab:col votes}
% shows the percentage of terms associated with the different colours.
Even though the colour options were presented in random order,
the order of the most frequently associated colours
is identical to the % order  % in which colours emerged in language 
% \cite{BerlinK69}
Berlin and Kay order
(Section 2:(1)).

The number of ambiguous words annotated was 2924.
1654 (57\%) of these words had senses that were associated with at least two different colours. 
Table \ref{tab:sensecolours} gives a few examples.

\begin{table}[t]
\centering
% \resizebox{\textwidth}{!}{
{\small
\begin{tabular}{l ll}
\hline
{\bf target}	&{\bf sense}	&{\bf colour}\\
\hline
bunk    &nonsense    &grey\\
bunk    &furniture   &brown\\
compatriot  &nation  &red\\
compatriot  &partner &white\\
frustrated  &hindrance   &red\\
frustrated  &disenchantment  &black\\
glimmer &idea    &white\\
glimmer &light   &yellow\\
stimulate   &allure  &red\\
stimulate   &encouragement   &green\\
\hline
\end{tabular}
}
\label{tab:sensecolours}
\vspace*{-2mm}
\caption{Example target words that have senses associated with different colours.}
\vspace*{2mm}
\end{table}

\begin{table}[t!]
% \caption{colour agreement.}
\centering
% \resizebox{0.5\textwidth}{!}{
 {\small
\begin{tabular}{rrrrrrr}
\hline
        \multicolumn{7}{c}{\bf majority class size} \\
                    one     &two    &three   &four   &five &$\geq$ two &$\geq$ three\\
\hline
        15.1        &52.9        &22.4        &7.3     &2.1   &84.9 &32.0  \\
\hline
\end{tabular}
 }
\label{tab:col agreement}
\vspace*{-3mm}
\caption{Percentage of terms in different majority classes.}
\vspace*{-4mm}
\end{table}

Table \ref{tab:col agreement} shows how often the majority class in colour associations
is 1, 2, 3, 4, and 5, respectively. 
% Since the annotators are given 11 different colour options to choose from, 
If we assume independence, then the chance that none of the 5 annotators agrees with each other
(majority class size of 1) is $1 \times 10/11 \times 9/11 \times 8/11 \times 7/11 = 0.344$.
Thus, if there was no correlation among any of the terms and colours, then 34.4\% of the time
none of the annotators would have agreed with each other. However, this happens only 15.1\% of the time.
A large number of terms have a majority class size $\geq$ 2 (84.9\%), and thus have more than chance association with a colour.
One can argue that terms with a majority class size $\geq$ 3 (32\%)  have {\it strong} colour associations.
% For about 2.1\% of the words, all 5 annotators agree on the associated colours.

 \begin{figure*}[t!]
  \begin{center}
  \includegraphics[width=1.5\columnwidth]{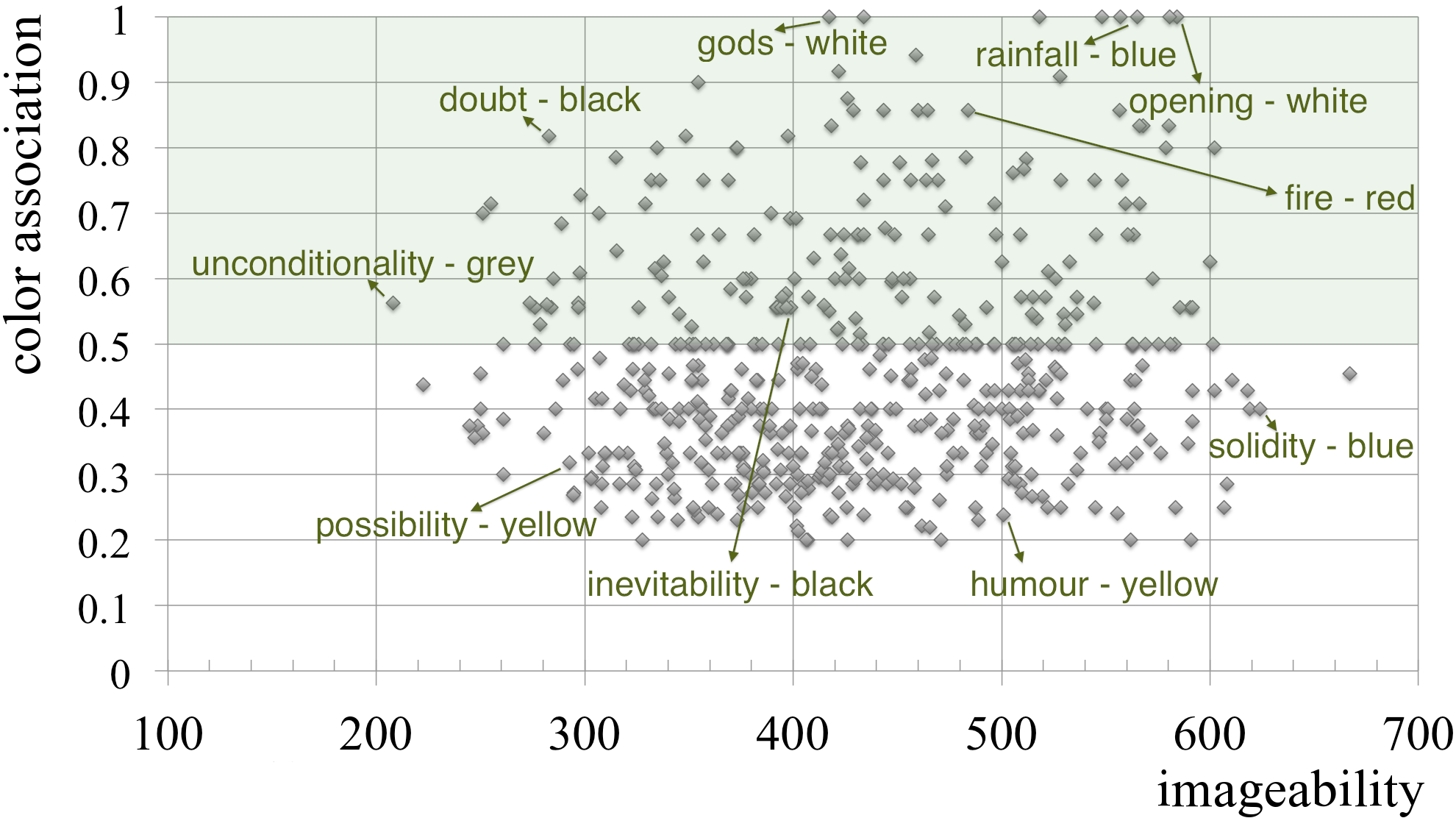}
  \end{center}
  % \caption{{\bf Polarity pie chart}---Proportions of polarities in quotations about gratitude. (Text from: Quote Garden)}
  % \caption{Scatter plot of imageability and association with colour of thesaurus categories.}
  \vspace*{-5mm}
  \caption{Scatter plot of thesaurus categories. The area of high colour association is shaded. Some points are labeled.}
  \label{fig:scatter}
  \vspace*{-3mm}
  \end{figure*}

Below are some reasons why agreement values are much lower than certain other tasks,
for example, part of speech tagging:
\begin{myitemize}
\item The annotators were not given a ``not associated with any colour" option.
Low agreement for certain instances is an indicator that these words have weak, if any, colour association.
% Also, this design makes kappa meaningless.
Therefore, inter-annotator agreement does not correlate with quality of annotation.
\item Words are associated with colours to different degrees.
Some words may be associated with more than one colour by comparable degrees, and there might be higher disagreement.
% , and there are no clear classes corresponding to different levels of association.
% Since we ask people to place term-colour associations in specific bins (no association and association), more people disagree for term--colour pairs
% whose degree of association is closer to the boundary, than for term--colour pairs further away from the boundary.
% \item \cite{Holsti69}, \cite{BrennanP81}, \cite{PerreaultL89}, and others consider
% the $\kappa$ values (both Fleiss's and Cohen's) to be conservative, especially
% when one category is much more prevalent than the other.
% In our data, the ``not associated with emotion" category is much more prevalent than the ``associated with emotion" category,
% so these $\kappa$ values might be underestimates of the true agreement.
\item The target word--sense pair is presented out of context. We expect higher agreement if
we provided words in context, but words can occur in innumerable contexts,
and annotating too many instances of the same word is costly.
% One can group contexts into those that correspond to different senses of the target word.
% By providing the word-choice question, we bias the Turker towards a particular sense
% of the target word, and aim to obtain the prior probability of the word sense's emotion association.
\end{myitemize}
\noindent Nonetheless, the lexicon is useful for downstream applications because
any of the following strategies may be employed: (1) choosing colour associations from only those instances
with high agreement, (2) assuming low-agreement terms have no colour association, (3) determining colour association of
a category through information from many words, as described in the next section.

\begin{table*}[t]
\centering
% \resizebox{\textwidth}{!}{
 {\small
\begin{tabular}{l rrrr rrrr rrr}
\hline
%        &\multicolumn{11}{c}{\bf colour} \\
            &white &black &red &green &yellow &blue &brown &pink &purple &orange &grey\\
\hline

anger words           &2.1    &{\bf 30.7}   &{\bf 32.4}   &5.0  &5.0  &2.4    &6.6    &0.5    &2.3    &2.5    &9.9\\
anticipation words    &{\bf 16.2}   &7.5    &11.5   &{\bf 16.2}   &10.7   &9.5    &5.7    &5.9    &3.1    &4.9    &8.4\\
disgust  words    &2.0  &{\bf 33.7}   &{\bf 24.9}   &4.8    &5.5    &1.9    &9.7    &1.1    &1.8    &3.5    &10.5\\
fear words        &4.5    &{\bf 31.8}   &{\bf 25.0} &3.5    &6.9    &3.0  &6.1    &1.3    &2.3    &3.3    &11.8\\
joy words     &{\bf 21.8}   &2.2    &7.4    &{\bf 14.1}   &13.4   &11.3   &3.1    &11.1   &6.3    &5.8    &2.8\\
sadness words     &3.0  &{\bf 36.0} &{\bf 18.6}   &3.4    &5.4    &5.8    &7.1    &0.5    &1.4    &2.1    &16.1\\
surprise  words       &11.0 &13.4   &{\bf 21.0} &8.3    &{\bf 13.5}   &5.2    &3.4    &5.2    &4.1    &5.6    &8.8\\
trust words       &{\bf 22.0} &6.3    &8.4    &14.2   &8.3    &{\bf 14.4}   &5.9    &5.5    &4.9    &3.8    &5.8\\
\hline
\end{tabular}
 }
\label{tab:col sig1}
\vspace*{-2mm}
\caption{Colour signature of emotive terms: percentage of terms associated with each colour.  For example, 32.4\% of the anger terms are associated with red.  The two most associated colours are shown in bold.}
\vspace*{1mm}
\end{table*}

% \vspace*{-14mm}
 \begin{table*}[t!]
 \centering
 % \resizebox{\textwidth}{!}{
  {\small
 \begin{tabular}{l rrrr rrrr rrr}
 \hline
%          &\multicolumn{11}{c}{\bf colour} \\
             &white &black &red &green &yellow &blue &brown &pink &purple &orange &grey\\
 \hline
 
 negative        &2.9    &{\bf 28.3}   &{\bf 21.6}   &4.7    &6.9    &4.1    &9.4    &1.2    &2.5    &3.8    &14.1\\
 positive        &{\bf 20.1}   &3.9    &8.0  &{\bf 15.5}   &10.8   &12.0 &4.8    &7.8    &5.7    &5.4    &5.7\\
 \hline
 \end{tabular}
  }
 \label{tab:col sig2}
\vspace*{-2mm}
 \caption{Colour signature of positive and negative terms: percentage terms associated with each colour.  For example, 28.3\% of the negative terms are associated with black.  The two most associated colours are shown in bold.}
\vspace*{-3mm}
 \end{table*}

\section{Category--Colour Association}

Different words within a thesaurus category may not be strongly associated with any colour, or they may be associated with many different colours.
We now determine whether there exist categories where the semantic coherence carries over to a strong common association with one colour. 

 We determine the strength of colour association of a category by first determining the colour $c$ most associated with the terms in it, 
 and then calculating the ratio of the number of times a word from the category is associated with $c$ to the number of words in the category associated with any colour. 
% The lowest possible strength of association is 1/11, and the highest is 1.
% Since our lexicon currently has entries for only about 10,500 words, 
% We calculated the strength of colour association of thesaurus categories using only those words appearing in in the word--colour lexicon. 
Only categories that had at least four words that also appear in the word--colour lexicon were considered; 535 of the 812  categories from  {\it Macquarie Thesaurus} met this condition. 
If a category has exactly four words that appear in the colour lexicon, and if all four words are associated with different colours, then the category has the lowest possible strength of colour association---0.25 (1/4). 
19 categories had a score of 0.25. No category had a score less than 0.25.
Any score above 0.25 shows more than random chance association with a colour. There were 516 such categories (96.5\%). 177 categories (33.1\%) had a score 0.5 or above, that is,
half or more of the words in these categories are associated with one colour. We consider these to be strong associations. 
% The existence of so many (blah \%) such categories gives further credence to the assertion that certain concepts indeed have a strong colour association.

\section{Imageability}

It is natural for physical entities of a certain colour to be associated with that colour. However, abstract concepts such as {\it danger} and {\it excitability} are also associated with colours---red and orange, respectively.\\
{\mbox Figure \ref{fig:scatter}} displays an experiment to determine whether there is a correlation between imageability and association with colour.

We define imageability of a thesaurus category to be the average of
the  imageability ratings of words in it.  We calculated imageability
for the 535 categories described in the previous section using
only the words that appear in the colour lexicon.  Figure
\ref{fig:scatter} shows the scatter plot of these categories on
the imageability and strength of colour association axes. 
% If there were a correlation between the two, that is 
If higher imageability correlated with
greater tendency to have a colour association, then we would 
see most of the points along the diagonal moving up from left to
right. Instead, we observe that the strongly associated categories % (points in the shaded region)  
are spread all across the imageability axis,
implying that there is only weak, if any, correlation.
% between imageability and strength of association with colour.
Imageability and colour association have a Pearson's product
moment correlation of 0.116, and a Spearman's rank order correlation of
0.102.

\section{The Colour of Emotion Words}
Emotions such as joy, sadness, and anger are abstract concepts dealing with one's psychological state.
As pointed out in Section 2, there is prior work on emotions evoked by colours.
In contrast, here we investigate the colours associated with emotion words.
We combine the word--emotion association lexicon compiled by Mohammad and Turney \shortcite{MohammadT10,MohammadT11} 
and our word--colour lexicon to
determine the colour signature of emotions---the rows in Table \ref{tab:col sig1}.
% The top two most frequently associated colours with each of the eight emotions are shown in bold.
Notably, we see that all of the emotions have strong associations with certain colours. 
% The ``anger" row shows the percentage of anger terms associated with different colours. 
Observe that anger is associated most with red.
Other negative emotions---disgust, fear, sadness---go strongest with black.
% However, for these emotions black is much more dominant. 
% Grey comes third for all four of the negative emotions.
Among the positive emotions: anticipation is most frequently associated with white and green;
joy with white, green, and yellow; and trust with white, blue, and green.
% Surprise, which can be positive or negative, is associated with red, yellow, and black.
Table \ref{tab:col sig2} shows the colour signature for terms marked positive and 
negative (these include terms that may not be associated
with the eight basic emotions).
% As expected from the results in Table \ref{tab:col sig1}, 
Observe that the negative terms are strongly associated with black and red, whereas the
positive terms are strongly associated with white and green.
Thus, colour can add to the potency of emotional concepts, yielding even more effective visualizations.
% Thus, we see that different emotions and different sentiments are strongly correlated with certain colours.

\section{Conclusions and Future Work}
We created a large word--colour association lexicon by crowdsourcing.
A word-choice question was used to guide the annotator to the desired sense of the target word, and to ensure data quality. % detect and discard erroneous input.
We observed that abstract concepts, emotions in particular, % and showed that they
have strong colour associations. % , especially emotions.
Thus, using the right colours in tasks such as information visualization, product marketing, and web development, can not only improve semantic coherence, but also
inspire the desired psychological response.
Interestingly, we found that frequencies of colour choice in associations follow the same order in which colour terms occur in language \cite{BerlinK69}.
% About 32\% of the words and blah\% of Roget thesaurus categories had a strong association with one of the eleven colours
% chosen for the experiment. We found no correlation between imageability and the tendency to have a colour association,
% but a strong association between emotions and colours. It was interesting to observe that the frequency with which colours were chosen
% in the association task reflects the order in which colours appeared in language.
Future work includes developing automatic corpus-based methods to determine the strength of word--colour association, and
% It is worth determining 
the extent to which strong word--colour associations manifest themselves as more-than-random chance co-occurrence in text.
% Occurrence of colour terms in text and associated images will be used to develop corpus- and image-based algorithms to determine colour associations.

\vspace*{-2mm}
{\small
\section*{Acknowledgments}
\vspace*{-3mm}
This research was funded by the National Research Council Canada (NRC).
Grateful thanks to Peter Turney, Tara Small, Bridget McInnes, and the reviewers for many wonderful ideas.
 Thanks to the more than 2000 people who answered the colour survey with diligence and care.
}

\bibliography{references}

\end{document}